\documentclass{llncs}
\usepackage{graphicx}
\usepackage{amsmath,amssymb} 
\usepackage{color}
\usepackage[width=122mm,left=12mm,paperwidth=146mm,height=193mm,top=12mm,paperheight=217mm]{geometry}
\usepackage{times}
\usepackage{epsfig}
\usepackage{multirow}
\usepackage{subfigure}
\usepackage{cellspace}
\usepackage[breaklinks=true,bookmarks=false]{hyperref}
\usepackage{moresize}
\usepackage{hyperref}
\usepackage{cite}
\begin{document}
\pagestyle{headings}
\mainmatter
\def\ECCV18SubNumber{****}  

\title{Part-Aligned Bilinear Representations \\for Person Re-identification}

\author{Yumin Suh$^\dagger$ \quad Jingdong Wang$^\ddagger$ \quad Siyu Tang$^{\ast \star}$ \quad Tao Mei$^\mathsection$ \quad Kyoung Mu Lee$^\dagger$}
\institute{
$^\dagger$ASRI, Seoul National University \quad
$^\ddagger$Microsoft Research\\
$^\ast$Max Planck Institute for Intelligent Systems\quad
$^\star$University of T\"{u}bingen\quad
$^\mathsection$JD.COM}

\maketitle

\begin{abstract}
We propose a novel network that learns a part-aligned representation for person re-identification. It handles the body part misalignment problem, that is, body parts are misaligned across human detections due to pose/viewpoint change and unreliable detection. Our model consists of a two-stream network (one stream for appearance map extraction and the other one for body part map extraction) and a bilinear-pooling layer that generates and spatially pools a part-aligned map. Each local feature of the part-aligned map is obtained by a bilinear mapping of the corresponding local appearance and body part descriptors. Our new representation leads to a robust image matching similarity, which is equivalent to an aggregation of the local similarities of the corresponding body parts combined with the weighted appearance similarity. This part-aligned representation reduces the part misalignment problem significantly. Our approach is also advantageous over other pose-guided representations (e.g., extracting representations over the bounding box of each body part) by learning part descriptors optimal for person re-identification. For training the network, our approach does not require any part annotation on the person re-identification dataset. Instead, we simply initialize the part sub-stream using a pre-trained sub-network of an existing pose estimation network, and train the whole network to minimize the re-identification loss. We validate the effectiveness of our approach by demonstrating its superiority over the state-of-the-art methods on the standard benchmark datasets, including Market-$1501$, CUHK$03$, CUHK$01$ and DukeMTMC, and standard video dataset MARS.
\keywords{Person re-identification}
\end{abstract}

\section{Introduction}
\label{sec:introduction}
The goal of person re-identification is to identify the same person across videos captured from different cameras, which is a fundamental visual recognition problem in video surveillance with various applications, including multi-people tracking~\cite{TangAAS17}. This process is challenging because the camera views are usually disjoint, the temporal transition time between cameras varies considerably, and the lighting conditions/person poses differ across cameras in real-world scenarios.

\begin{figure}[t]
\centering
    \begin{minipage}{0.99\linewidth}
\centering
      \includegraphics[height=0.18\linewidth]{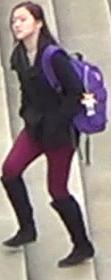}
      \includegraphics[height=0.18\linewidth]{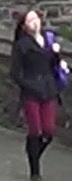}
      \includegraphics[height=0.18\linewidth]{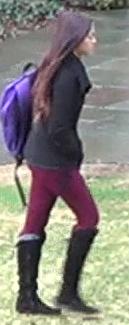}
      \includegraphics[height=0.18\linewidth]{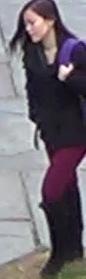}
      \includegraphics[height=0.18\linewidth]{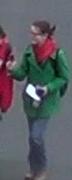}
      \includegraphics[height=0.18\linewidth]{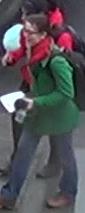}
      \includegraphics[height=0.18\linewidth]{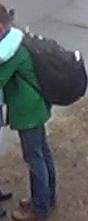}
      \includegraphics[height=0.18\linewidth]{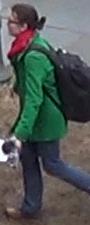}
      \includegraphics[height=0.18\linewidth]{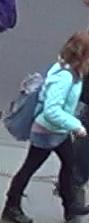}
      \includegraphics[height=0.18\linewidth]{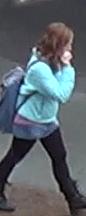}
      \includegraphics[height=0.18\linewidth]{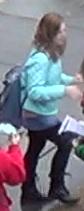}
      \includegraphics[height=0.18\linewidth]{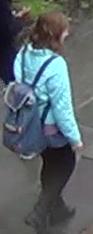}
    \end{minipage}

    \begin{minipage}{0.99\linewidth}
    \begin{minipage}{0.33\linewidth}
\centering
      \small{(a)}
    \end{minipage}
    \begin{minipage}{0.33\linewidth}
\centering
      \small{(b)}
    \end{minipage}
    \begin{minipage}{0.33\linewidth}
\centering
      \small{(c)}
    \end{minipage}
    \end{minipage}
\vspace{-.3cm}

\caption{(a, b) As a person appears in different poses/viewpoints in different cameras, and (c) human detections are imperfect, the corresponding body parts are usually not spatially aligned across the human detections, causing person re-identification to be challenging.}
\label{fig:main}
\end{figure}

Body part misalignment (i.e., the problem that body parts are spatially misaligned across person images) is one of the key challenges in person re-identification. Figure~\ref{fig:main} shows some examples. This problem causes conventional global representations~\cite{PaisitkriangkraiSV15,XiaoLOW16,ZhengBSWSWT16,PersonNet2016,chen2016deep,xiaoli2016end} or strip/grid-based representations~\cite{deepreid2014,improved2015,DCSL2016ijcai,deepmetric2014,ChengGZWZ16,VariorSLXW16} to be unreliable as they implicitly assume that every person appears in a similar pose within a tightly surrounded bounding box. Thus, a body part-aligned representation, which can ease the representation comparison and avoid the need of complex comparison techniques, should be designed.

To resolve this problem, recent approaches have attempted to localize body parts explicitly and combine the representations over them~\cite{SuLZXGT17,ZhengHLY17,ZhaoTSSYYWT17,conf/cvpr/li17,ZhaoLZW17}. For example, the body parts are represented by the pre-defined (or refined~\cite{SuLZXGT17}) bounding boxes estimated from the state-of-the-art pose estimators~\cite{ZhengHLY17,SuLZXGT17,conf/cvpr/cao17,ZhaoTSSYYWT17}. This scheme requires highly-accurate pose estimation. Unfortunately, state-of-the-art pose estimation solutions still probably fail in person re-identification datasets.
In addition, these schemes are bounding box-based and lack fine-grained part localization within the boxes.
Recently, Zhao et al.~\cite{ZhaoLZW17} represented body parts through confidence maps, which are estimated using attention techniques. The method has a lack of guidance on body part locations during the training, thereby failing to consistently attend to certain body regions.

In this paper, we propose a novel body part-aligned representation to address the misalignment problem. Our approach learns to represent the human poses as part maps and combine them directly with the appearance maps to compute part-aligned representations, and circumvents the aforementioned problems. More precisely, our model consists of a two-stream network and an aggregation module. 1) Each stream separately generates appearance and body part maps. 2) The aggregation module first generates the part-aligned feature maps by computing the bilinear mapping of the appearance and part descriptors at each location, and then spatially averages the local part-aligned descriptors. For the training of the network, we do not use any body part annotations on the person re-identification dataset. Instead, we simply initialize the part map generation stream using the pre-trained weights, which are trained from a standard pose estimation dataset. Surprisingly, although our approach only optimizes the re-identification loss function, the resulting two-stream network successfully separates appearance and part information into each stream, thereby generating the appearance and part maps from each of them, respectively. {\color{black}The part maps adapt to differentiate/blur informative/unreliable body parts} and further support person re-identification. We show how the similarity between two images computed over our part-aligned representation is different from and superior to that computed over previous body-part box-based representations. Through extensive experiments, we verify that our approach consistently improves the accuracy of the baseline and achieves competitive/superior performance over standard image datasets, Market-$1501$, CUHK$03$,  CUHK$01$ and DukeMTMC, and one standard video dataset, MARS.

\section{Related Work}
The early solutions of person re-identification mainly relied on hand-crafted features~\cite{ma2012local,LOMO2015,viewpoint2008,MatsukawaOSS16}, metric learning techniques~\cite{null2016,ZhangLLIR16,PaisitkriangkraiSV15,psd2015,KodirovXFG16,LiZWXG15,JingZWYKYHX15}, and probabilistic patch matching algorithms~\cite{dapeng2015,ChenYCZ16,correspondence2015} to handle resolution/light/view/pose changes. Recently, attributes~\cite{SuYZTDLG15,SuZX0T16,ZhaoOW14}, transfer learning~\cite{PengXWPGHT16,ShiHX15}, re-ranking~\cite{ZhengWTHLT15,GarciaMMG15},
partial person matching~\cite{ZhengLXLLG15}, and human-in-the-loop learning~\cite{MartinelDMR16,WangGZX16},
have also been studied. More can be found from the survey~\cite{ZhengYH16}.
In the following,
we review recent spatial-partition-based and part-aligned representations, matching techniques, 
and some works using bilinear pooling.

\vspace{0.1cm}
\noindent\textbf{Regular spatial-partition based representations.}
The approaches in this stream of research represent an image as a combination of local descriptors, where each local descriptor represents a spatial partition such as grid cell~\cite{deepreid2014,improved2015,DCSL2016ijcai} and horizontal stripe~\cite{deepmetric2014,ChengGZWZ16,VariorSLXW16}. They work well under a strict assumption that the location of a certain body part is consistent across images. This assumption is often violated under realistic conditions, thereby causing the methods to fail. 
An extreme case is that no spatial partition is used and a global representation is computed over the whole image~\cite{PaisitkriangkraiSV15,XiaoLOW16,ZhengBSWSWT16,PersonNet2016,chen2016deep,xiaoli2016end}.

\vspace{0.1cm}
\noindent\textbf{Body part-aligned representations.}
Body part and pose detection results were ever exploited for person re-identification to handle the body part misalignment problem~\cite{ChengCSBM11,XuLZL13,BakCBT10a,Farenzena2010symmetry,WeinrichVG13,ChengC14}. Recently, these ideas have been re-studied using deep learning techniques. Most approaches~\cite{ZhengHLY17,SuLZXGT17,ZhaoTSSYYWT17} represent an image 
as a combination of body part descriptors, where a dozen of pre-defined body parts are detected using the off-the-shelf pose estimator (possibly an additional RoI refinement step). They usually crop bounding boxes around the detected body parts and compute the representations over the cropped boxes. The difference from our approach is that we use part maps, where the part is not predefined
and learned from the person re-identification problem in the form of a spatial map instead of cropped boxes.

Tang et al~\cite{TangAAS17} also introduced part maps for person re-identification to solve the multi-people tracking problem. They used part maps to augment appearances as another feature, rather than to generate part-aligned representations, which is different from our method. Some works~\cite{LiuZTSSYYW17,ZhaoLZW17} proposed the use of attention maps, which are expected to attend to informative body parts. They often fail to produce reliable attentions as the attention maps are estimated from the appearance maps; guidance from body part locations is lacking, resulting in a limited performance.

\vspace{0.1cm}
\noindent\textbf{Matching.}
The simple similarity functions~\cite{deepmetric2014,VariorSLXW16,ChengGZWZ16}, e.g., cosine similarity or Euclidean distance,
are adapted, for part-aligned representations, such as our approach, or under an assumption that the representations are body part/pose aligned. Various schemes~\cite{WangZLZZ16,improved2015,deepreid2014,DCSL2016ijcai} are designed to eliminate the influence
from body part misalignment for spatial partition-based representations. For instance, a matching sub-network is proposed to conduct convolution and max-pooling operations, over the differences~\cite{improved2015} or the concatenation~\cite{deepreid2014,DCSL2016ijcai} of grid-based representation of a pair of person images. Varior et al.~\cite{VariorHW16} proposed the use of matching maps in the intermediate features to guide feature extraction in the later layers through a gated CNN. 

\vspace{0.1cm}
\noindent\textbf{Bilinear pooling.}
Bilinear pooling is a scheme to aggregate two different types of feature maps by using the outer product at each location and spatial pooling them to obtain a global descriptor. This strategy has been widely adopted in fine-grained recognition~\cite{conf/iccv/lin15,conf/cvpr/gao16,conf/iclr/kim17} and shows promising performance. For person re-identification, Ustinova et al.~\cite{conf/avss/ustinova17} adopted a bilinear pooling to aggregate two different appearance maps; this method does not generate part-aligned representations and leads to poor performance. 
Our approach uses a bilinear pooling to aggregate appearance and part maps to compute part-aligned representations.

\begin{figure*}[t]
  \begin{minipage}{0.29\linewidth}
   \end{minipage}
  \begin{minipage}{0.89\linewidth}
   \centering
		\includegraphics[width=.99\linewidth]{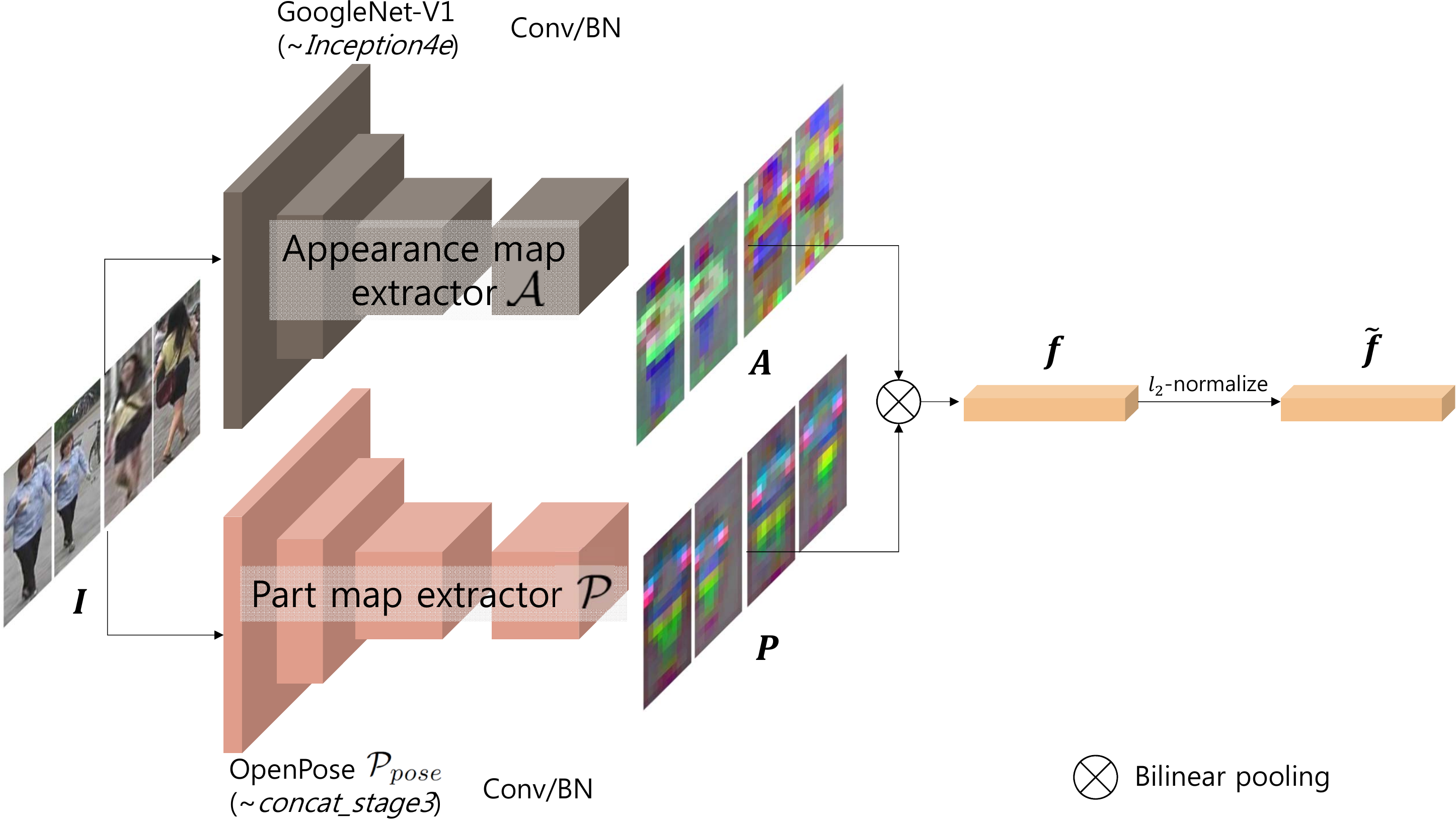}
\end{minipage}
\vspace{-0.1cm}
\caption{Overview of the proposed model. The model mainly consists of a two-stream network and an aggregator (bilinear pooling). For a given image $\mathbf{I}$, the appearance and part map extractors, $\mathcal{A}$ and $\mathcal{P}$, generate the appearance and part maps, $\mathbf{A}$ and $\mathbf{P}$, respectively. The aggregator performs bilinear pooling over $\mathbf{A}$ and $\mathbf{P}$ and generates a feature vector $\mathbf{f}$. Finally, the feature vector is $l_2$-normalized, resulting in a final part-aligned representation $\mathbf{\tilde{f}}$. {\color{black}Conv and BN denote the convolution and batch normalization layers, respectively.}}\label{fig:pipeline}
\end{figure*}

\section{Our Approach}
The proposed model consists of a two-stream network and an aggregation module. It receives an image $\mathbf{I}$ as an input and outputs a part-aligned feature representation $\tilde{\mathbf{f}}$ as illustrated in Figure~\ref{fig:pipeline}. The two-stream network contains two separate sub-networks, the appearance map extractor $\mathcal{A}$ and the part map extractor $\mathcal{P}$, which extracts the appearance map $\mathbf{A}$ and part map \textbf{P}, respectively. The two types of maps are aggregated through bilinear pooling to generate the part-aligned feature $\mathbf{f}$ and subsequently normalized to generate the final feature vector $\tilde{\mathbf{f}}$.

\subsection{Two-Stream Network}
\noindent\textbf{Appearance map extractor.}
We feed an input image $\mathbf{I}$ into the
appearance map extractor $\mathcal{A}$, thereby
outputting the appearance map
$\mathbf{A}$:
\begin{align}
\mathbf{A} = \mathcal{A}(\mathbf{I}).
\label{eq:appearance_map}
\end{align}
$\mathbf{A} \in \mathbb{R}^{h \times w \times c_A}$ is a feature map of size $h \times w$, where each location is described by $c_A$-dimensional local appearance descriptor. We use the sub-network of GoogLeNet~\cite{inceptionv1} to form and initialize $\mathcal{A}$.

\noindent\textbf{Part map extractor.}
The part map extractor $\mathcal{P}$ receives an input image $\mathbf{I}$ and outputs the part map $\mathbf{P}$:
\begin{align}
\mathbf{P}=\mathcal{P}(\mathbf{I}).
\end{align}
$\mathbf{P}\in \mathbb{R}^{h\times w \times c_P}$ is a feature map of size $h \times w$, where each location is described by a $c_P$-dimensional local part descriptor. 
Considering the rapid progress in pose estimation, we use the sub-network of the pose estimation network, OpenPose~\cite{conf/cvpr/cao17}, to form and initialize $\mathcal{P}$. We denote the sub-network of the OpenPose as $\mathcal{P}_{pose}$.

\subsection{Bilinear Pooling}
Let $\mathbf{a}_{xy}$ be the appearance descriptor at the position $(x,y)$ from the appearance map $\mathbf{A}$, and $\mathbf{p}_{xy}$ be the part descriptor at the position $(x,y)$ from the part map $\mathbf{P}$.
We perform bilinear pooling over $\mathbf{A}$ and $\mathbf{P}$ to compute the part-aligned representation $\mathbf{f}$.
There are two steps,
bilinear transformation
and spatial global pooling,
which are mathematically given as follows:
\begin{align}
\mathbf{f} = 
\operatorname{pooling}_{xy}\{\mathbf{f}_{xy}\}
= \frac{1}{S} \sum_{xy}\mathbf{f}_{xy},
\quad\quad
\mathbf{f}_{xy} = \operatorname{vec} (\mathbf{a}_{xy}\otimes \mathbf{p}_{xy}),
\label{eqn:bilinearpooling1}
\end{align}
where $S$ is the spatial size. The pooling operation we use here is average-pooling. 
$\operatorname{vec}(.)$ transforms a matrix to a vector,
and
$\otimes$ represents the outer product of two vectors,
with the output being a matrix. 
The part-aligned feature $\mathbf{f}$ is then normalized to generate the final feature vector $\tilde{\mathbf{f}}$ as follows:
\begin{align}
\tilde{\mathbf{f}} = \frac{\mathbf{f}}{\|\mathbf{f} \|_2}.
\label{eqn:bilinearpooling2}
\end{align}
Considering the normalization, we denote the normalized part-aligned representation as
$\tilde{\mathbf{f}}_{xy} = 
\operatorname{vec}(\tilde{\mathbf{a}}_{xy}\otimes \tilde{\mathbf{p}}_{xy})$, where $\tilde{\mathbf{a}}_{xy} = \frac{\mathbf{a}_{xy}}{\sqrt{\|\mathbf{f}\|_2}}$ and  $\tilde{\mathbf{p}}_{xy} = \frac{\mathbf{p}_{xy}}{\sqrt{\|\mathbf{f}\|_2}}${\color{black}. Therefore, $\tilde{\mathbf{f}} = \frac{1}{S}\sum_{xy}{\tilde{\mathbf{f}}_{xy}}$.}

\vspace{.1cm}
\noindent\textbf{Part-aligned interpretation.}
We can decompose $\mathbf{a} \otimes \mathbf{p}$\footnote{We drop the subscript $xy$
for presentation clarification}
into $c_P$ components:
\begin{align}
\operatorname{vec}(\mathbf{a} \otimes \mathbf{p})
=
[(p_1\mathbf{a})^\top~
(p_2\mathbf{a})^\top~ 
\dots
(p_{c_P}\mathbf{a})^\top]^\top,
\label{eq:pose_aligned}
\end{align}
where each sub-vector $p_i \mathbf{a}$ corresponds to a $i$-th part channel. For example, if $p_{knee}=1$ on knee and $0$ otherwise, then $p_{knee} \mathbf{a}$ becomes $\mathbf{a}$ only on the knee and $\mathbf{0}$ otherwise. Thus, we call $\operatorname{vec}(\mathbf{a}\otimes\mathbf{p})$ as part-aligned representation. In general, each channel $c$ does not necessarily correspond to a certain body part. However, the part-aligned representation remains valid as $\mathbf{p}$ encodes the body part information. Section {\ref{section:analysis}} describes this interpretation in detail.

\subsection{Loss}
To train the network, we utilize the widely-used triplet loss function.
Let $\mathbf{I}_q$, $\mathbf{I}_p$ and $\mathbf{I}_n$ denote the query, positive and negative images, respectively. Then, $(\mathbf{I}_q, \mathbf{I}_p)$ is a pair of images of the same person, and $(\mathbf{I}_q, \mathbf{I}_n)$ is that of different persons.
Let $\tilde{\mathbf{f}}_q$, $\tilde{\mathbf{f}}_p$, and $\tilde{\mathbf{f}}_n$ indicate their representations. The triplet loss function is formulated as
\begin{equation}
\label{eq:triplet_loss}
\ell_{\operatorname{triplet}}(\tilde{\mathbf{f}}_q,\tilde{\mathbf{f}}_p,\tilde{\mathbf{f}}_n)
=  \max(m + \operatorname{sim}(\tilde{\mathbf{f}}_q,\tilde{\mathbf{f}}_n) - \operatorname{sim}(\tilde{\mathbf{f}}_q,\tilde{\mathbf{f}}_p), 0),
\end{equation}
where $m$ denotes a margin and
$\operatorname{sim}(\mathbf{x}, \mathbf{y}) = <\mathbf{x}, \mathbf{y}>$. 
The margin is empirically set as
$m = 0.2$.
The overall loss function is written as follows.
\begin{align}
\mathcal{L} = \frac{1}{|\mathcal{T}|}\sum\nolimits_{(\mathbf{I}_q, \mathbf{I}_p, \mathbf{I}_n) \in
\mathcal{T}}\ell_{\operatorname{triplet}}(\tilde{\mathbf{f}}_q,\tilde{\mathbf{f}}_p,\tilde{\mathbf{f}}_n),
\label{eqn:overalllossfunction}
\end{align}
where $\mathcal{T}$ is the set
of all triplets, $\{(\mathbf{I}_q, \mathbf{I}_p, \mathbf{I}_n)\}$.

\section{Analysis}
\label{section:analysis}

\noindent\textbf{Part-aware image similarity.}
We show that under the proposed part-aligned representation in Eqs.(\ref{eqn:bilinearpooling1}) and (\ref{eqn:bilinearpooling2}), the similarity between two images is equivalent to the aggregation of local appearance similarities between the corresponding body parts. 
The similarity between two images can be represented as the sum of local similarities between every pair of locations as follows.
\begin{align}
\label{eq:img_sim}
\operatorname{sim}_I(\mathbf{I}, \mathbf{I}')
=<\tilde{\mathbf{f}}, \tilde{\mathbf{f}}'> \nonumber
=&\frac{1}{S^2}<\sum_{xy}\tilde{\mathbf{f}}_{xy}, \sum_{x'y'}\tilde{\mathbf{f}}'_{x'y'}> \nonumber \\
=&\frac{1}{S^2}\sum_{xy} \sum_{x'y'} <\tilde{\mathbf{f}}_{xy} , \tilde{\mathbf{f}}'_{x'y'} > \nonumber \\
= & \frac{1}{S^2}\sum_{xy} \sum_{x'y'}  \operatorname{sim}( \tilde{\mathbf{f}}_{xy}, \tilde{\mathbf{f}}'_{x'y'}),
\end{align}
where $\operatorname{sim}_I(,)$ measures the similarity between images.
Here, the local similarity is computed by an inner product:
\begin{align}
\label{eq:local_sim}
\operatorname{sim}(\tilde{\mathbf{f}}_{xy}, \tilde{\mathbf{f}}_{x'y'}')
&= <\operatorname{vec}(\tilde{\mathbf{a}}_{xy} \otimes \tilde{\mathbf{p}}_{xy}),
\operatorname{vec}(\tilde{\mathbf{a}}'_{x'y'}  \otimes\tilde{\mathbf{p}}'_{x'y'}) > \nonumber \\
&=<\tilde{\mathbf{a}}_{xy}, \tilde{\mathbf{a}}'_{x'y'}>
 <\tilde{\mathbf{p}}_{xy},\tilde{\mathbf{p}}'_{x'y'}> \nonumber \\
 &= \operatorname{sim}(\tilde{\mathbf{a}}_{xy}, \tilde{\mathbf{a}}'_{x'y'})
 \operatorname{sim}(\tilde{\mathbf{p}}_{xy}, \tilde{\mathbf{p}}'_{x'y'}).
\end{align}
This local similarity can be interpreted as the appearance similarity weighted by the body part similarity or vice versa.
Thus, from Eqs(\ref{eq:img_sim}) and (\ref{eq:local_sim}), the similarity between two images is computed as the average of local appearance similarities weighted by the body part similarities at the corresponding positions:
\begin{equation}
\operatorname{sim}_I(\mathbf{I}, \mathbf{I}') = \frac{1}{S^2}\sum_{xyx'y'}{ \operatorname{sim}(\tilde{\mathbf{a}}_{xy}, \tilde{\mathbf{a}}'_{x'y'})
 \operatorname{sim}(\tilde{\mathbf{p}}_{xy}, \tilde{\mathbf{p}}'_{x'y'})}. \nonumber
\end{equation}

\noindent\textbf{Relationship to the baseline models.} Consider a baseline approach that only uses the appearance maps and spatial global pooling for image representation. Then, the image similarity is computed as $\operatorname{sim}_I(\mathbf{I}, \mathbf{I}') = \frac{1}{S^2}\sum_{xyx'y'}{ \operatorname{sim}(\tilde{\mathbf{a}}_{xy}, \tilde{\mathbf{a}}'_{x'y'})}$.
Unlike our model, this approach cannot reflect part similarity.
Consider another model based on the box-based representation, which represents an image as a concatenation of $K$ body part descriptors, where $k$-th body part is represented as the average-pooled appearance feature within the corresponding bounding box. This model is equivalent to our model when $\mathbf{p}_{xy}$ is defined as $\mathbf{p}_{xy}=[\delta[(x,y)\in R_{1}], \cdots, \delta[(x,y)\in R_{K}]]$, where $R_k$ is the region within the $k$-th part bounding box and $\delta[\cdot]$ is an indicator function, i.e., $\delta[x] = 1$ if $x$ is true otherwise $0$.
Because our model contains these baselines as special cases and is trained to optimize the re-identification loss, it is guaranteed to perform better than them.

\begin{figure*}[t]
\centering  
	\begin{minipage}{0.99\linewidth}
	\includegraphics[width=0.99\linewidth]{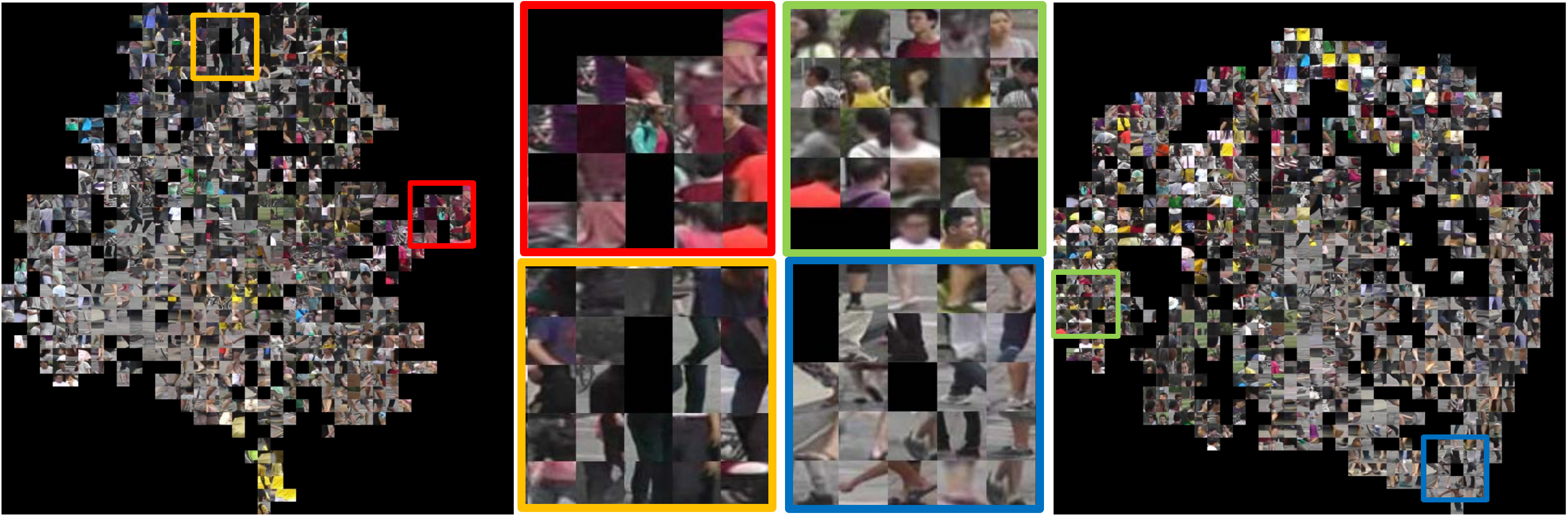} 
	\end{minipage}
\vspace{.2cm}

	\begin{minipage}{0.49\linewidth}
	\centering
	\small(a) Appearance features
	\end{minipage}
	\begin{minipage}{0.49\linewidth}
	\centering
	\small(b) Part features
	\end{minipage}
\vspace{-.3cm}
\caption{The t-SNE visualization of the normalized local appearance and part descriptors on the Market-1501 dataset. It illustrates that our two-stream network decomposes the appearance and part information into two streams successfully. (a) Appearance descriptors are clustered roughly by colors, independently from the body parts where they came from. (b) Part descriptors are clustered by body parts where they came from, regardless of the colors.
(Best viewed on a monitor when zoomed in)
} \label{fig:tsne}
\label{fig:separatemaps}
\end{figure*}

\begin{figure}[t]
\centering
	\begin{minipage}{0.99\linewidth}
	\begin{minipage}{0.24\linewidth}
\centering
      \includegraphics[width=0.99\linewidth]{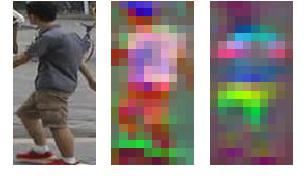}
      \includegraphics[width=0.99\linewidth]{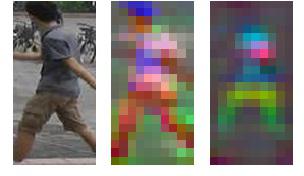}
      \includegraphics[width=0.99\linewidth]{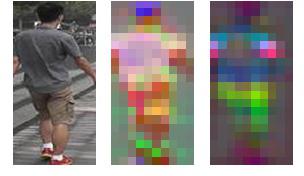}
      \includegraphics[width=0.99\linewidth]{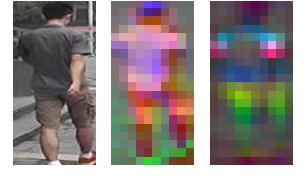}
    \end{minipage}
	\begin{minipage}{0.24\linewidth}
\centering
      \includegraphics[width=0.99\linewidth]{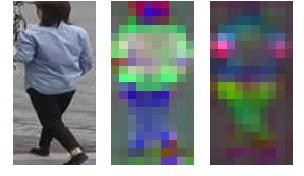}
      \includegraphics[width=0.99\linewidth]{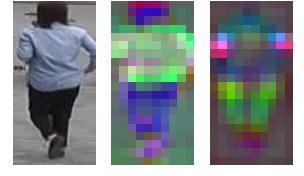}
      \includegraphics[width=0.99\linewidth]{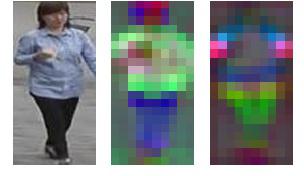}
      \includegraphics[width=0.99\linewidth]{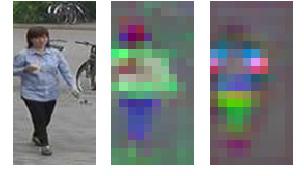}
    \end{minipage}
	\begin{minipage}{0.24\linewidth}
\centering
      \includegraphics[width=0.99\linewidth]{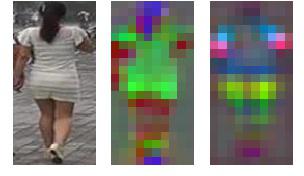}
      \includegraphics[width=0.99\linewidth]{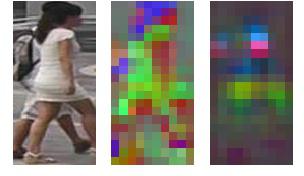}
      \includegraphics[width=0.99\linewidth]{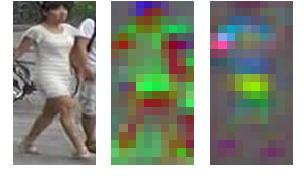}
      \includegraphics[width=0.99\linewidth]{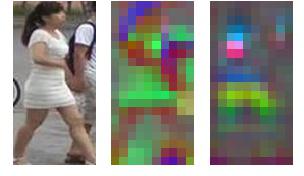}
    \end{minipage}
	\begin{minipage}{0.24\linewidth}
\centering
      \includegraphics[width=0.99\linewidth]{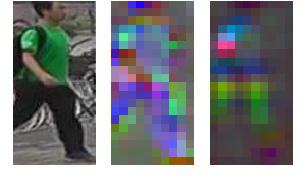}
      \includegraphics[width=0.99\linewidth]{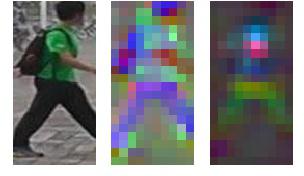}
      \includegraphics[width=0.99\linewidth]{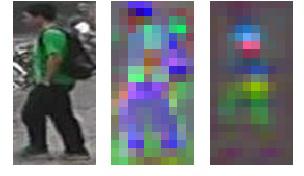}
      \includegraphics[width=0.99\linewidth]{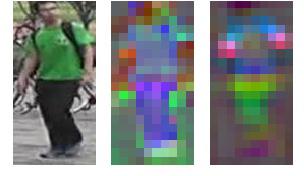}
    \end{minipage}

    \end{minipage}
\vspace{-.2cm}
\caption{Visualization of the appearance maps $\mathbf{A}$ and part maps $\mathbf{P}$ obtained from the proposed method. For a given input image (left), appearance (center) and part (right) maps encode the appearance and body parts, respectively. For both appearance and part maps, the same color implies that the descriptors are similar, whereas different colors indicate that the descriptors are different. The appearance maps share similar color patterns among the images from the same person, which means that the patterns of appearance descriptors are similar as well. In the part maps, the color differs depending on the location of the body parts where the descriptors came from. (Best viewed in color)}
\label{fig:vis}
\vspace{-.1cm}
\end{figure}

\begin{figure}[t]
     \begin{minipage}{0.99\linewidth}
\centering
      \includegraphics[width=0.24\linewidth]{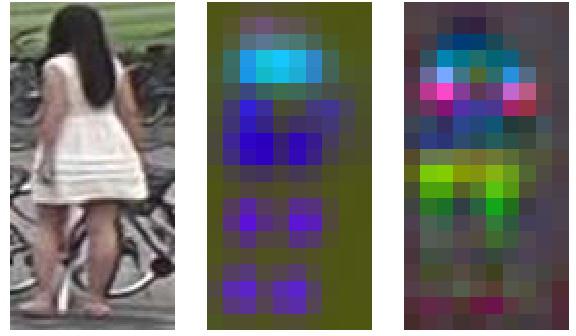}
      \includegraphics[width=0.24\linewidth]{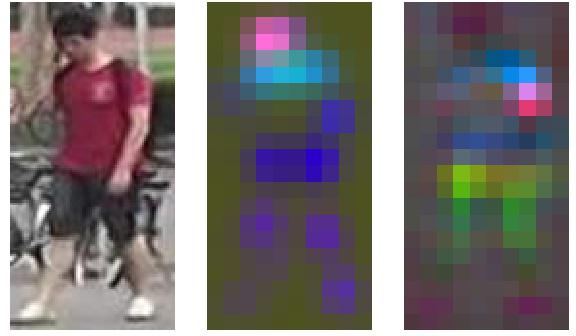}
      \includegraphics[width=0.24\linewidth]{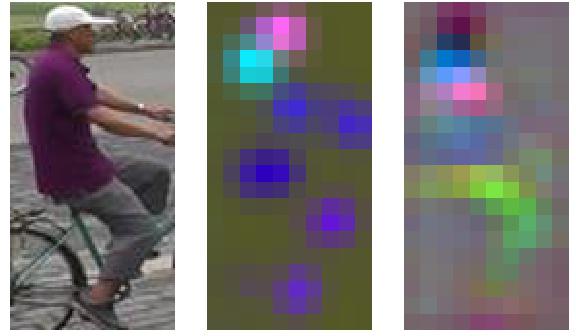}
      \includegraphics[width=0.24\linewidth]{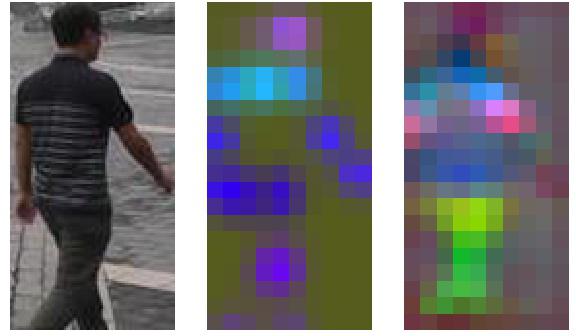}
    \end{minipage}
\vspace{-.1cm}
\caption{Comparing the body part descriptors. For a given image (left), the conventional joint-based (center) and the proposed (right) descriptors are visualized. (Best viewed in color)}
\label{fig:posemap_compare}
\vspace{-.5cm}
\end{figure}

\vspace{.1cm}
\noindent\textbf{The two-stream network yields a decomposed appearance and part maps.}
At the beginning of the training, the two streams of the network {\color{black}mainly represent the appearance and part maps because the appearance map extractor $\mathcal{A}$ and the part map extractor $\mathcal{P}$ are initialized using GoogleNet~\cite{googlenet} pre-trained on ImageNet~\cite{imagenet} and OpenPose~\cite{conf/cvpr/cao17} model pre-trained on COCO~\cite{coco}, respectively.}
During training, we do not set any constraints on the two streams, i.e., no annotations for the body parts, but only optimize the re-identification loss. Surprisingly, the trained two-stream network maintains to decompose the appearance and part information into two streams: one stream corresponds to the appearance maps and the other corresponds to the body part maps.

We visualize the distribution of the learned local appearance and part descriptors using t-SNE~\cite{tsne}
~\footnote{\url{https://cs.stanford.edu/people/karpathy/cnnembed}} as shown in Figures~\ref{fig:tsne} (a) and (b). Figure~\ref{fig:tsne} (a) shows that the appearance descriptors are clustered depending on the appearance while being independent on the parts that they come from. For example, the red/yellow box shows that the red/black-colored patches are closely embedded, respectively. By contrast, Figure~\ref{fig:tsne} (b) illustrates that the local part embedding maps the similar body parts into close regions regardless of color. For example, the green/blue box shows that the features from the head/lower leg are clustered, respectively. In addition, physically adjacent body parts, such as head--shoulder and shoulder--torso, are also closely embedded.

To understand how the learned appearance/part descriptors are used in person re-identification, we visualize the appearance maps $\mathbf{A}$ and the part maps $\mathbf{P}$ following the visualization used in SIFTFlow~\cite{siftflow}, as shown in Figure~\ref{fig:vis}.~\footnote{we project the $c_A$(or $c_P$)-dimensional local descriptor vector onto the 3D RGB space, by mapping the top three principal components of the descriptor to the principal components of RGB.} For a given input image (left), the appearance (center) and part (right) maps encode the appearance and body parts, respectively. The figure shows how the appearance maps differentiate different persons while being invariant for each person.
By contrast, the part maps encode the body parts independently from their appearance. In particular, a certain body part is represented by similar color across images, which confirms our observation in Figure~\ref{fig:tsne} that the part features from physically adjacent regions are closely embedded.

Our approach learns the optimal part descriptor for person re-identification, rather than relying on the pre-defined body parts. Figure~\ref{fig:posemap_compare} qualitatively compares the conventional body part descriptor and the one learned by our approach.~\footnote{We used the visualization method proposed in SIFTFlow~\cite{siftflow}} In the previous works on human pose estimation~\cite{conf/cvpr/wei16,conf/cvpr/cao17,hourglass}, human poses are represented as a collection of pre-defined key body joint locations. It corresponds to a part descriptor which one-hot encodes the key body joints depending on the existence of a certain body joint at the location, e.g, $p_{knee}=1$ on knee and $0$ otherwise. 
Compared to the baseline, ours smoothly maps the body parts. In other words, the colors are continuous over the whole body in ours, which implies that the adjacent body parts are mapped closely. By contrast, the baseline not always maps adjacent body parts maps closely.  For example, the upper leg between the hip and knee is more close to the background descriptors than to ankle or knee descriptors. This smooth mapping renders our method to work robustly against the pose estimation error because the descriptors do not change rapidly along the body parts and therefore are insensitive to the error in estimation. In addition, the part descriptors adopt to more finely distinguish the informative parts. For example, the mapped color varies sharply from elbow to shoulder and differentiates the detailed regions. Based on these properties, the learned part descriptors better support the person re-identification task and improve the accuracy.

\section{Implementation Details}
\label{section:implementation_details}
\noindent\textbf{Network architecture.}
We use a sub-network of the first version of GoogLeNet~\cite{googlenet} as the appearance map extractor $\mathcal{A}$, 
from the image input of size $160 \times 80$ to the output of \textit{inception4e}, which is followed by a $1\times 1$ convolution layer and a batch normalization layer to reduce the dimension to $512$ (Figure~\ref{fig:pipeline}). Moreover, we optionally adopt dilation filters in the layers from the \textit{inception4a} to the final layer, resulting in $20 \times 10$ response maps. Figure~\ref{fig:pipeline} illustrates the architecture of the part map extractor $\mathcal{P}$. We use a sub-network of the OpenPose network~\cite{conf/cvpr/cao17}, from the image input to the output of stage2 (i.e., \textit{concat\_stage3}) to extract $185$ pose heat maps, which is followed by a $3\times 3$ convolution layer and a batch normalization layer, thereby outputting $128$ part maps. 
We adopt the compact bilinear pooling~\cite{conf/cvpr/gao16} to aggregate the two feature maps into a $512$-dimensional vector $\mathbf{f}$.

\vspace{0.1cm}
\noindent\textbf{Compact bilinear pooling.}
The bilinear transformation
over the $512$-dimensional appearance vector
and the $128$-dimensional part vector
results in an extremely high dimensional vector,
which consumes large computational cost and memory.
To resolve this issue, we use the tensor sketch approach~\cite{TensorSketch}
to compute a compact representation
as in~\cite{conf/cvpr/gao16}.
The key idea of the tensor sketch approach is that
the original inner product,
on which the Euclidean distance is based,
between two high-dimensional vectors can be approximated
as an inner product of the dimension-reduced vectors,
which are random projections of the original vectors.
Details can be found in \cite{TensorSketch}.

\vspace{0.1cm}
\noindent\textbf{Network training}.
The appearance map extractor $\mathcal{A}$ and part map extractor $\mathcal{P}$ are fine-tuned from the network pre-trained on ImageNet~\cite{imagenet} and COCO~\cite{coco}, respectively. The added layers are initialized following~\cite{xavier}.
We use the stochastic gradient descent algorithm.
The initial learning rate, weight decay, and the momentum are set to $0.01$, $2 \times 10^{-3}$, and $0.9$, respectively.
The learning rate is decreased by a factor of $5$ after every $20,000$ iterations.
All the networks are trained for $75,000$ iterations.

We follow~\cite{ZhaoLZW17} to sample a mini-batch of samples
at each iteration
and use all the possible triplets within each mini-batch.
The gradients are computed using the acceleration trick
presented in~\cite{ZhaoLZW17}.
In each iteration, we sample a mini-batch of $180$ images, e.g., there are on average $18$ identities with each containing 10 images. In total, there are approximately $10^2\cdot(180-10)\cdot 18 \approx3 \times10^5$ triplets in each iteration.

\vspace{0.1cm}
\noindent\textbf{Pose sub-network $\mathcal{P}_{pose}$.}
Table~\ref{table:jointlimb} compares the accuracy when different pose sub-networks $\mathcal{P}_{pose}$ are used. {\it  joint\_only, limb\_only, internal only}, and {\it joint\_limb\_internal} (proposed) denotes a network that generates the joint-based, limb-based, and internal confidence maps with $19$, $38$, $128$, and $185$ channels, respectively, from OpenPose~\cite{conf/cvpr/cao17}.
{\color{black}It shows that the proposed method achieves similar accuracy for joints and limbs. As the internal feature map provides complementary information to the joints/limbs, we use their concatenation in the final model.}

\begin{table*}[t]
\setlength{\tabcolsep}{5pt}
\caption{\small Accuracy comparison on various pose sub-networks $\mathcal{P}_{pose}$ on Market-$1501$}\vspace{-.3cm}
\label{table:jointlimb}
\centering
\scriptsize
\begin{tabular}[pos]{|Sl| |Sc|Sc|Sc|Sc|Sc|}
\hline
Rank & $1$ & $5$ & $10$ & $20$ &mAP \\
\hline
{\it joint\_only} & $88.9$ & $96.0$ & $97.3$ & $98.3$ & $75.6$\\
{\it limb\_only} & $\mathbf{90.5}$ & $95.9$ & $97.3$ & $98.0$ & $75.5$\\
{\it internal\_only} & $88.2$ & $95.4$ & $97.1$ & $98.1$ & $74.3$\\
{\it joint\_limb\_internal} (proposed) & $90.2$ & $\mathbf{96.1}$ & $\mathbf{97.4}$ & $\mathbf{98.4}$ & $\mathbf{76.0}$\\
\hline
\end{tabular}\vspace{-.4cm}
\end{table*}

\section{Experiments}
\subsection{Datasets}

\noindent
{\bf Market-$1501$}~\cite{conf/iccv/zheng15}. This dataset is one of the largest benchmark datasets for person re-identification.
Six cameras are used: five high-resolution cameras and one low-resolution camera.
There are $32,668$ DPM-detected pedestrian image boxes of $1,501$ identities:
$750$ identities are utilized for training and
the remaining $751$ identities are used for testing.
There are $3,368$ query images and $19,732$ gallery images
with $2,793$ distractors.

\vspace{0.1cm}
\noindent{\bf CUHK$03$}~\cite{deepreid2014}. This dataset consists of
$13,164$ images of $1,360$ people
captured by six cameras.
Each identity appears in two disjoint camera views (i.e., $4.8$ images in each view on average).
We divided the train/test set following the previous work~\cite{deepreid2014}.
For each test identity, two images are randomly sampled as the probe and gallery images and the average accuracy over $20$ trials is reported as the final result.

\vspace{0.1cm}
\noindent{\bf CUHK$01$}~\cite{conf/accv/li12}. This dataset comprises $3884$ images of
$971$ people captured in two disjoint camera views.
Two images are captured for each person from each of the two cameras (i.e., a total of four images).
Experiments are performed under two evaluation settings~\cite{improved2015}, using $100$ and $486$ test IDs.
Following the previous works~\cite{improved2015,chen2016deep,ChengGZWZ16,ZhaoLZW17},
we fine-tuned the model from the one learned from the CUHK$03$ training set for the experiments with $486$ test IDs.

\vspace{0.1cm}
\noindent{\bf DukeMTMC}~\cite{DukeMTMC}. This dataset is originally proposed for video-based person tracking and re-identification.
We use the fixed train/test split and evaluation setting following \cite{journal/arxiv/lin17}\footnote{\url{https://github.com/layumi/DukeMTMC-reID_evaluation}}.
It includes $16,522$ training images of $702$ identities,
$2,228$ query images of $702$ identities and $17,661$ galley images.

\vspace{0.1cm}
\noindent{\bf MARS}~\cite{ZhengBSWSWT16}. This dataset is proposed for video-based person re-identification.
It consists of $1261$ different pedestrians captured by at least two cameras.
There are $509,914$ bounding boxes and $8,298$ tracklets from $625$ identities for training
and $681,089$ bounding boxes and $12,180$ tracklets from $636$ identities for testing.

\subsection{Evaluation Metrics}
We use both the cumulative matching characteristics (CMC) and mean average precision
(mAP) to evaluate the accuracy.
The CMC score measures the quality of identifying the correct match at each rank.
For multiple ground truth matches, CMC cannot measure how well all the images are ranked.
Therefore, we report the mAP scores for Market-$1501$, DukeMTMC, and MARS
where more than one ground truth images are in the gallery.

\begin{figure} [t]
\centering

\small
\includegraphics[width=0.32\linewidth]{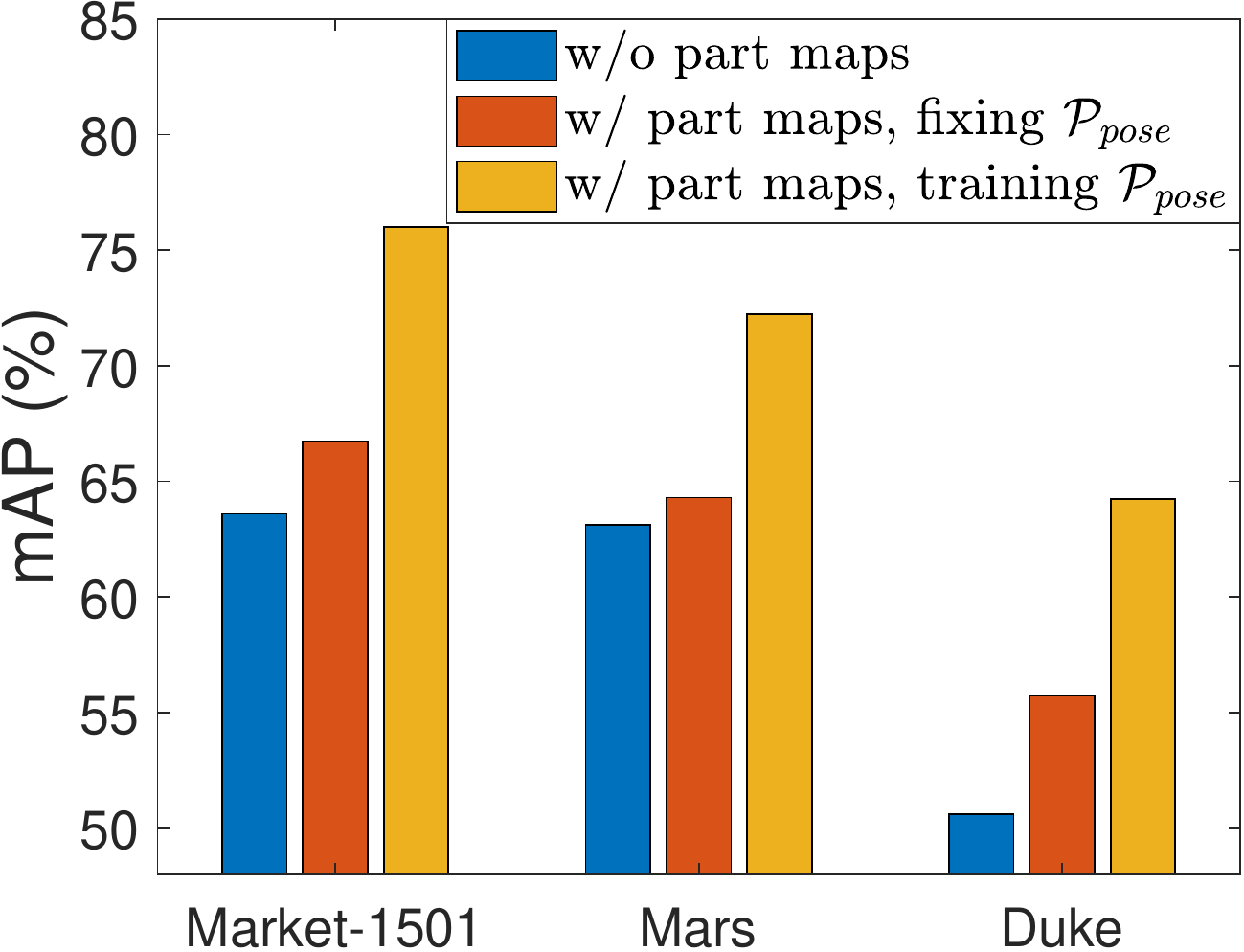}
\includegraphics[width=0.32\linewidth]{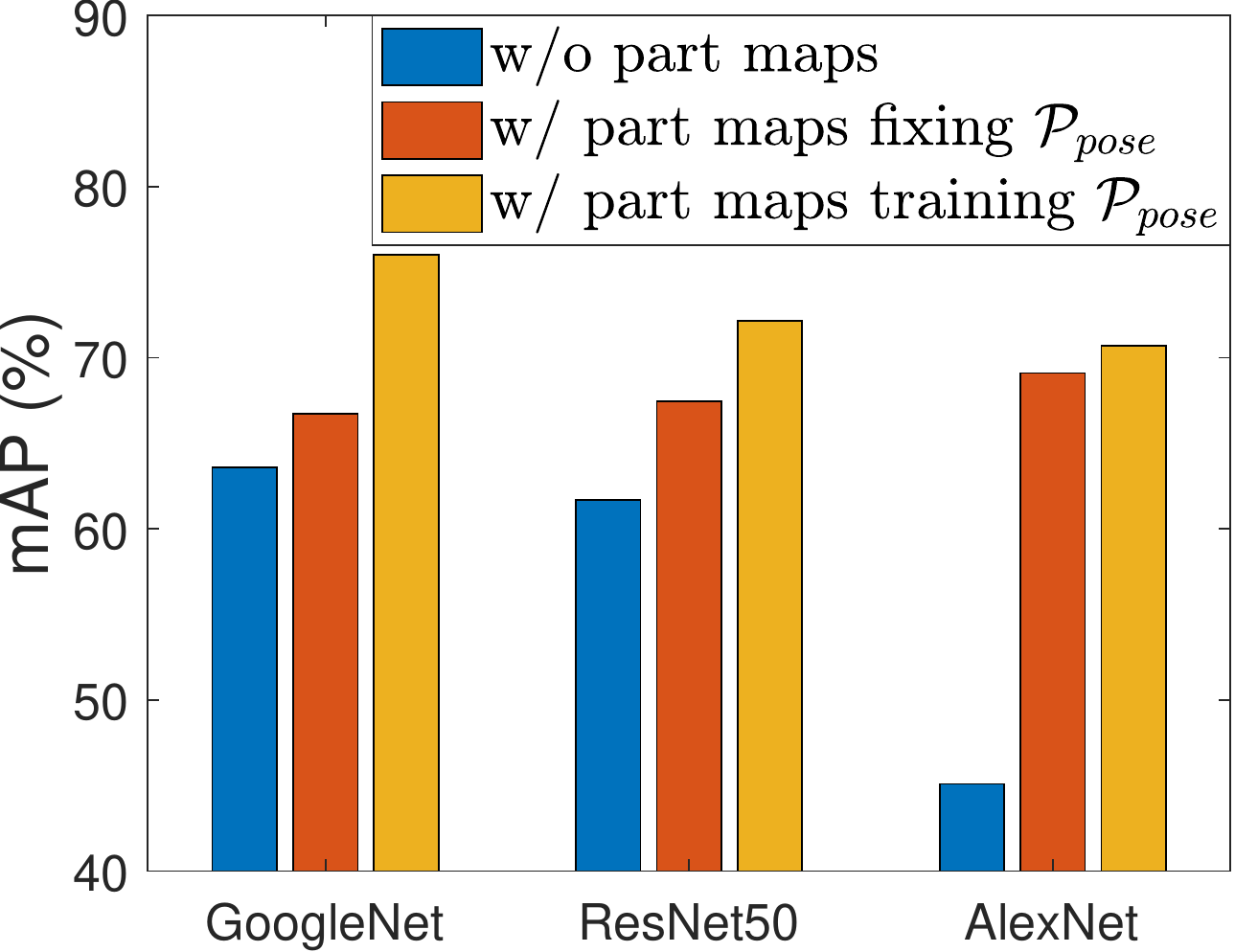}
\includegraphics[width=0.32\linewidth]{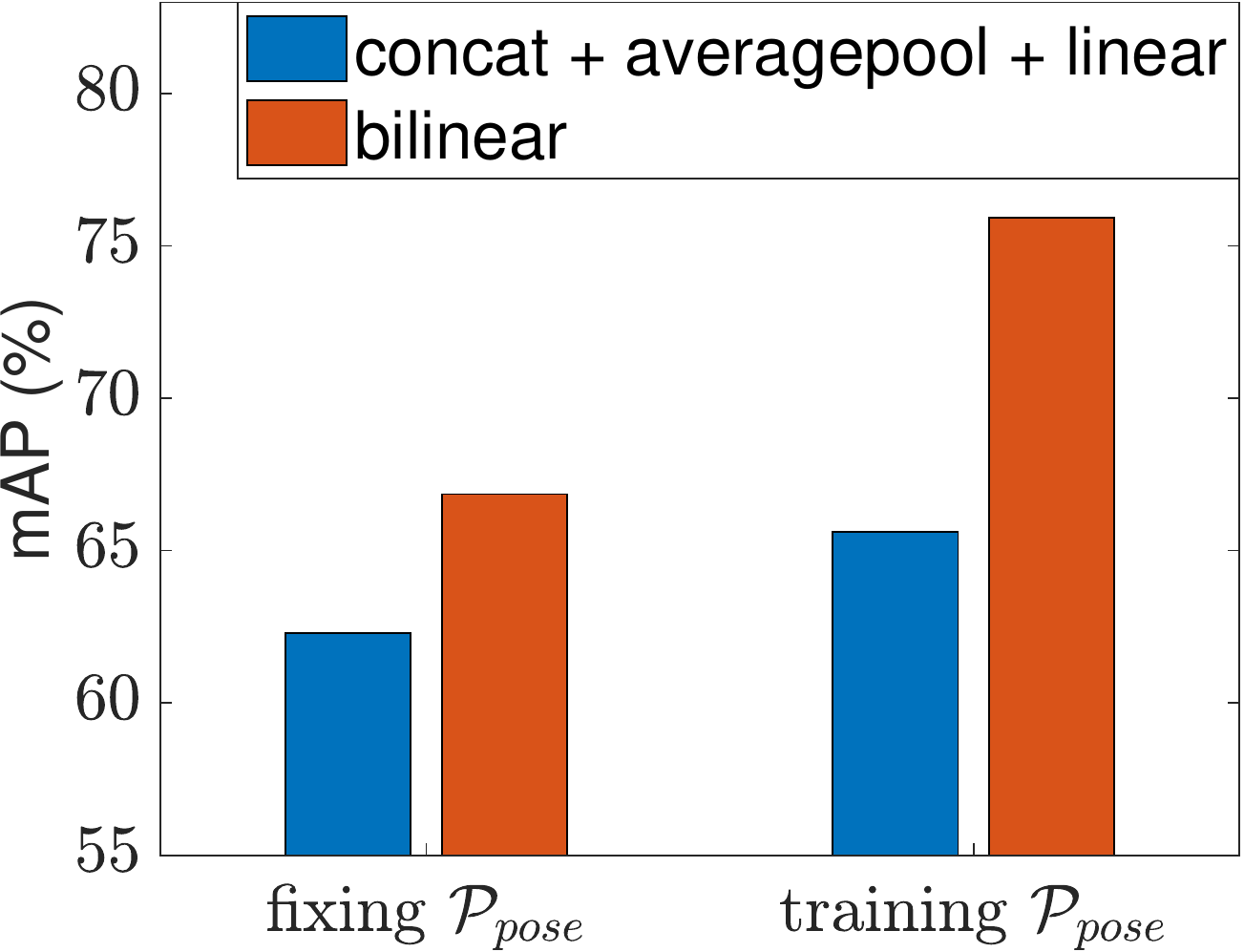}

    \begin{minipage}{0.32\linewidth}
\centering
      \small{(a)}
    \end{minipage}
    \begin{minipage}{0.32\linewidth}
\centering
      \small{(b)}
    \end{minipage}
    \begin{minipage}{0.32\linewidth}
\centering
      \small{(c)}
    \end{minipage}
\vspace{-.3cm}
\caption{
(a) Comparison of different pooling methods on the appearance maps. (c) Comparing models, with and without part maps, on different datasets. (d) Comparing models, with and without part maps, on different architectures of the appearance map extractor. If not specified, the results are reported on Market-$1501$. (b) Comparison of different methods to aggregate the appearance and part maps.
}
\label{fig:baseline}
\vspace{-.5cm}
\end{figure}

\subsection{Comparison with the Baselines}
\label{section:comparison_with_baselines}
We compare the proposed method with the baselines in three aspects. In this section, when not specified, all the experiments are performed on the Market-1501 dataset, all the models do not use dilation, and $\mathcal{P}_{pose}$ is trained together with the other parameters.

\noindent\textbf{Effect of part maps.}
We compare our method with a baseline that does not explicitly use body parts.
As a baseline network, we use the appearance map extractor of Eq.(\ref{eq:appearance_map}), which is followed by a global spatial average pooling and a fully connected layer, thereby outputting the $512$-dimensional image descriptor.
Figures~\ref{fig:baseline} (a) and (b) compare the proposed method with the baseline, while varying the training strategy: fixing and training $\mathcal{P}_{pose}$. Fixing $\mathcal{P}_{pose}$ initializes $\mathcal{P}_{pose}$ using the pre-trained weights~\cite{conf/cvpr/cao17,coco} and fixes the weight through the training. Training $\mathcal{P}_{pose}$ also initializes $\mathcal{P}_{pose}$ in the same way, but fine-tunes the network using the loss of Eq.(\ref{eqn:overalllossfunction}) during training.
Figure~\ref{fig:baseline} (a) illustrates the accuracy comparison on three datasets,  Market-$1501$, MARS, and Duke. It shows that using part maps consistently improves the accuracy on all the three datasets from the baseline. In addition, training $\mathcal{P}_{pose}$ largely improves the accuracy than fixing $\mathcal{P}_{pose}$. It implies that the part descriptors are adopted to better serve the person re-identification task. Figure~\ref{fig:baseline} (b) shows the accuracy comparison while varying the {\color{black} appearance sub-network architecture}. Similarly, the baseline accuracy is improved when part maps are introduced and further improved when $\mathcal{P}_{pose}$ is fine-tuned during training.

\noindent\textbf{Effect of bilinear pooling.}
Figure~\ref{fig:baseline} (c) compares the proposed method ({\it bilinear}) to the baseline with a different aggregator. For the given appearance and part maps, {\it concat+averagepool+linear} generates a feature vector by concatenating two feature maps, spatially average pooling, and feeding through a fully connected layer, resulting in a $512$-dimensional vector. The result shows that bilinear pooling consistently achieves higher accuracy than the baseline, for both cases when $\mathcal{P}_{pose}$ is fixed/trained.

\noindent\textbf{Comparison with previous pose-based methods.}
Finally, we compare our method with three previous works~\cite{ZhengHLY17,ZhaoTSSYYWT17,SuLZXGT17}, which use human pose estimation, on Market-1501. For a fair comparison, we use the reduced CPM(R-CPM [$\sim$3M param]) utilized in~\cite{ZhaoTSSYYWT17}\footnote{\url{https://github.com/yokattame/SpindleNet}} as $\mathcal{P}_{pose}$. The complexity of the R-CPM is lower than the standard FCN ($\sim$6M param) used in \cite{SuLZXGT17} and CPM ($\sim$30M param) used in \cite{ZhengHLY17}. As the appearance network, \cite{SuLZXGT17} used GoogLeNet and \cite{ZhengHLY17} used ResNet50. \cite{ZhaoTSSYYWT17} used 13 inception modules, whereas we use 7. Table~\ref{table:result_market} shows the comparison. In comparison with the method adopted by \cite{ZhengHLY17,ZhaoTSSYYWT17,SuLZXGT17}, the proposed method (Inception V1, R-CPM) achieves an increase of $4$\% and $9$\% for rank@1 accuracy and mAP, respectively. It shows that our method effectively uses the part information compared with the previous approaches.

\subsection{Comparison with State-of-the-Art Methods}

\noindent\textbf{Market-$1501$}.
Table~\ref{table:result_market} shows the comparison over two query schemes, single query and multi-query. Single query takes one image from each person whereas multi-query takes multiple images. For the multi-query setting, one descriptor is obtained from multiple images by averaging the feature from each image. Our approach achieves the best accuracy in terms of both mAP and rank@K for both single and multi-query.
We also provide the result after re-ranking~\cite{conf/cvpr/zhong17}, which further boosts accuracy.  In addition, we conduct the experiment over an expanded dataset with additional $500K$ images~\cite{conf/iccv/zheng15}. Following the standard evaluation protocol~\cite{journal/arxiv/hermans17}, we report the results over four different gallery sets, 
$19, 732$,
$119,732$,
$219,732$,
and $519,732$,
using two evaluation metrics (i.e., rank-$1$ accuracy and mAP). Table~\ref{table:result_market_500k} reports the results. The proposed method outperforms all the other methods.

\vspace{.1cm}
\noindent\textbf{CUHK$03$}.
We report the results
with two person boxes:
manually labeled
and detected.
Table~\ref{table:result_cuhk03} presents the comparison with existing solutions.
In the case of detected boxes, the state-of-the-art accuracy is achieved. With manual bounding boxes, our method also achieves the best accuracy.

\begin{table*}
\setlength{\tabcolsep}{4pt}
\caption{\small Accuracy comparison on Market-$1501$}
\label{table:result_market}
\centering
\scriptsize
\begin{tabular}{|Sl|| Sc|Sc|Sc|Sc|Sc|  Sc|Sc|Sc|Sc|Sc|}
\hline
 & \multicolumn{5}{c|}{Sinlge Query} & \multicolumn{5}{c|}{Multi Query} \\
\hline
Rank & 1 & 5 & 10 & 20 & mAP & 1 & 5 & 10 & 20 & mAP \\
\hline\hline
Varior et al. 2016~\cite{VariorHW16} & $61.6$ & - & - & - & $35.3$ & -& -& -& -& -\\
Zhong et al. 2017~\cite{conf/cvpr/zhong17} & $77.1$ & - & - & - & $63.6$  & -& -& -& -& -\\
Zhao et al. 2017~\cite{ZhaoLZW17} & $80.9$ & $91.7$ & $94.7$ & $96.6$ & $63.4$ & -& -& -& -& -\\
Sun et al. 2017~\cite{conf/iccv/sun17} & $82.3$ & $92.3$ & $95.2$ & - & $62.1$ & - & -& -& -& -\\
Geng et al. 2016~\cite{journal/arxiv/geng16} & $83.7$ & - & - & - & $65.5$ & $89.6$ & -& -& -& $73.8$\\
Lin et al. 2017~\cite{journal/arxiv/lin17} & $84.3$ & $93.2$ & $95.2$ & $97.0$ & $64.7$ & -& -& -& -& -\\
Bai et al. 2017~\cite{journal/arxiv/bai17} & $82.2$ & - & - & - & $68.8$ & $88.2$ & -& -& -& $76.2$\\
Chen et al. 2017~\cite{journal/pami/chen17} & $72.3$ & $88.2$ & $91.9$ & $95.0$ & - & -& -& -& -&\\
Hermans et al. 2017~\cite{journal/arxiv/hermans17} & $84.9$ & $94.2$ & - &- & $69.1$  & $90.5$ & $96.3$ & -& -& $76.4$\\
\quad\quad{\color{blue}+ re-ranking} & {\color{blue} $86.7$}& {\color{blue}$93.4$} & {\color{blue}-} &{\color{blue}-} & {\color{blue}$81.1$} & {\color{blue}$91.8$} & {\color{blue}$95.8$} & - & - & {\color{blue} $87.2$}\\
Zhang et al. 2017~\cite{journal/arxiv/zhang17} & $87.7$ & - & - &- & $68.8$  & $91.7$ & -& -& -& $77.1$\\
Zhong et al. 2017~\cite{journal/arxiv/zhong17} & $87.1$ & - & - &- & $71.3$  & - & -& -& -& -\\
\quad\quad{\color{blue}+ re-ranking} & {\color{blue}$89.1$}& - & - & - &{\color{blue} $83.9$} & - & - & - & - & -\\
Chen et al. 2017~\cite{conf/cvpr/chen17} (MobileNet) & $90.0$ & - & - & - & $ 70.6$ & - & - & - & - & -\\
Chen et al. 2017~\cite{conf/cvpr/chen17} (Inception-V3) & $88.6$ & - & - & - & $72.6$ & - & - & - & - & -\\
\hline
Ustinova et al. 2017 \cite{conf/avss/ustinova17} (Bilinear) & $66.4$ & $85.0$ & $90.2$ & - & $41.2$  & - & -& -& -& -\\
\hline
Zheng et al. 2017~\cite{ZhengHLY17} (Pose) & $79.3$ & $90.8$ & $94.4$ & $96.5$ & $56.0$  & -& -& -& -&-\\
Zhao et al. 2017~\cite{ZhaoTSSYYWT17} (Pose) & $76.9$ & $91.5$ & $94.6$ & $96.7$ & - & -& -& -& -& -\\
Su et al. 2017~\cite{SuLZXGT17} (Pose) & $84.1$ & $92.7$ & $94.9$ & $96.8$ & $65.4$ & - & - & - & - & -\\
\hline
Proposed (Inception-V1, R-CPM) &  $88.8$ & $95.6$ & $97.3$ &  $98.6$ & $74.5$ & $92.9$ & $97.3$ & $98.4$ & $99.1$ & $81.7$\\
Proposed (Inception-V1, OpenPose)& $\mathbf{90.2}$ & $\mathbf{96.1}$ & $\mathbf{97.4}$ & $\mathbf{98.4}$ & $\mathbf{76.0}$ & $\mathbf{93.2}$ & $\mathbf{97.5}$ & $\mathbf{98.4}$ & $\mathbf{99.1}$ & $\mathbf{82.7}$ \\
\quad\quad + dilation& $\mathbf{91.7}$ & $\mathbf{96.9}$ & $\mathbf{98.1}$ & $\mathbf{98.9}$ & $\mathbf{79.6}$ & $\mathbf{94.0}$ & $\mathbf{98.0}$ & $\mathbf{98.8}$ & $\mathbf{99.3}$ & $\mathbf{85.2}$\\
\quad\quad{\color{blue}+ re-ranking} & {\color{blue} $\mathbf{93.4}$} & {\color{blue} $\mathbf{96.4}$} & {\color{blue}$\mathbf{97.4}$} & {\color{blue} $\mathbf{98.2}$} & {\color{blue} $\mathbf{89.9}$} & {\color{blue} $\mathbf{95.4}$} &{\color{blue} $\mathbf{97.5}$} & {\color{blue} $\mathbf{98.2}$} & {\color{blue} $\mathbf{98.9}$} & {\color{blue} $\mathbf{93.1}$}\\
\hline
\end{tabular}\vspace{-.0cm}
\end{table*}

\begin{table}\centering
\setlength{\tabcolsep}{5pt}
\caption{\small Accuracy comparison on Market-$1501$+$500$k.}
\label{table:result_market_500k}
\resizebox{0.70\linewidth}{!}
{
\begin{tabular}{|Sl|| Sc|Sc|Sc|Sc|Sc|}
\hline
 & \multicolumn{5}{c|}{Gallery size} \\
\hline
 & metric & $19732$ & $119732$ & $219732$ & $519732$ \\
\hline\hline
\multirow{2}{*}{Zheng et al. 2017~\cite{journal/arxiv/zheng16b}} & rank-$1$ & $79.5$ & $73.8$ & $71.5$ & $68.3$ \\
 & mAP & $59.9$ & $52.3$ & $49.1$ & $45.2$ \\
\hline
\multirow{2}{*}{Linet al. 2017~\cite{journal/arxiv/lin17}} & rank-$1$ & $84.0$ & $79.9$ & $78.2$ & $75.4$ \\
 & mAP & $62.8$ & $56.5$ & $53.6$ & $49.8$ \\
\hline
\multirow{2}{*}{Hermans et al. 2017~\cite{journal/arxiv/hermans17}} & rank-$1$ &$ 84.9$ & $79.7$ & $77.9$ & $74.7$ \\
 & mAP & $69.1$ & $61.9$ & $58.7$ & $53.6$ \\
\hline
\multirow{2}{*}{Proposed (Inception V1, OpenPose)} & rank-1 & $\mathbf{91.7}$ & $\mathbf {88.3}$ & $\mathbf{86.6}$ & $\mathbf{84.1}$\\
 & mAP & $\mathbf{79.6}$ & $\mathbf{74.2}$ & $\mathbf{71.5}$ & $\mathbf{67.2}$\\

\hline
\end{tabular}
}\vspace{-.0cm}
\end{table}

\begin{table*}
\setlength{\tabcolsep}{2pt}
\caption{\small Accuracy comparison on CUHK$03$ and CUHK$01$}
\label{table:result_cuhk03}
\centering
\tiny
\begin{tabular}[pos]{|Sl| |Sc|Sc|Sc|Sc| |Sc|Sc|Sc|Sc|  |Sc|Sc|Sc|Sc| |Sc|Sc|Sc|Sc|}
\hline
 & \multicolumn{8}{c|}{CUHK$03$} & \multicolumn{8}{c|}{CUHK$01$} \\
\hline
 & \multicolumn{4}{c|}{Detected} & \multicolumn{4}{c|}{Manual}& \multicolumn{4}{c|}{$100$ test IDs} & \multicolumn{4}{c|}{$486$ test IDs} \\
\hline
Rank & $1$ & $5$ & $10$ & $20$ & $1$ & $5$ & $10$ & $20$ & $1$ & $5$ & $10$ & $20$ & $1$ & $5$ & $10$ & $20$\\
\hline\hline
Shi et al.~\cite{deepmetric2014} & $52.1$ & $84.0$ & $92.0$ & $96.8$ & $61.3$ & $88.5$ & $96.0$ & $99.0$ & $69.4$ & $90.8$ & $96.0$ & - &- &- &- &-\\
SIR-CIR ~\cite{WangZLZZ16} & $52.2$ &-&-&-&-&-&-&-& $71.8$ & $91.6$ & $96.0$ & $98.0$ &- &- &- &- \\
Varior et al.~\cite{VariorHW16} & $68.1$ & $88.1$ & $94.6$ & $98.8$ &-&-&-&-& -& -& -& -& -& -& -& -\\
Bai et al.~\cite{journal/arxiv/bai17} &$72.7$ & $92.4$ & $96.1$ & - &$76.6$ & $94.6$ & $98.0$ & - &-& -& -& -& -& -& -& -\\
Zhang et al.~\cite{DCSL2016ijcai}& -&-&-&-&$80.2$&$97.7$&$99.2$&$99.8$& $89.6$ & $97.8$ & $98.9$ & $99.7$ & $76.5$ & $94.2$ & $\mathbf{97.5}$  & - \\
Sun et al.~\cite{conf/iccv/sun17} &$81.8$ & $95.2$ & $97.2$ & - & - & - & - & -&-& -& -& -& -& -& -& -\\
Zhao et al.~\cite{ZhaoLZW17} & $81.6$ & $97.3$ & $98.4$ & $\mathbf{99.5}$ & $85.4$ & $97.6$ & $99.4$ & $\mathbf{99.9}$& $88.5$ & $\mathbf{98.4}$ & $\mathbf{99.6}$ &$\mathbf{99.9}$ & $74.7$ & $92.6$ & $96.2$  & $98.4$ \\
Geng et al.~\cite{journal/arxiv/geng16} & $84.1$ & - & - & -  & $85.4$ & - & - & -  & $\mathbf{93.2}$ &-&-&-& $77.0$&-&-&- \\
Chen et al.~\cite{journal/pami/chen17} & $87.5$ & $97.4$ & $98.7$ & $\mathbf{99.5}$& -& -& -& - &-&-&-&-&$74.5$ & $91.2$ & $94.8$ & $97.1$
\\
\hline
Ustinova et al.~\cite{conf/avss/ustinova17} (Bilinear) & $63.7$ & $89.2$ & $94.7$ & $97.5$ & $69.7$ & $93.4$ & $98.9$ & $99.4$   & -& -& -& - &  $52.9$ & $78.1$ & $86.3$ & $92.6$\\
\hline
Zheng et al.~\cite{ZhengHLY17} (Pose) & $67.1$ & $92.2$ & $96.6$ & $98.1$  &-&-&-&- &-&-&-&- &-&-&-&-\\
Zhao et al.~\cite{ZhaoTSSYYWT17} (Pose) & -& -& -& - & $88.5$ & $97.8$ & $98.6$ & $99.2$ & -& -& -& - & $79.9$ & $\mathbf{94.4}$ & $97.1$ & $98.6$\\
Su et al.~\cite{SuLZXGT17} (Pose) & $78.3$ & $94.8$ & $97.2$ & $98.4$ & $88.7$ & $98.6$ & $99.2$ & $99.7$&-&-&-&- &-&-&-&-\\
\hline
Proposed & $\mathbf{88.0}$ & $\mathbf{97.6}$ & $\mathbf{98.6}$ & $\mathbf{99.0}$ & $\mathbf{91.5}$ & $\mathbf{99.0}$ & $\mathbf{99.5}$ & $\mathbf{99.9}$ &$90.4$& $97.1$ & $98.1$ & $98.9$ &  $\mathbf{80.7}$ & $\mathbf{94.4}$ & $\mathbf{97.3}$ & $\mathbf{98.6}$ \\
\hline
\end{tabular}\vspace{-.0cm}
\end{table*}

\vspace{.1cm}
\noindent\textbf{CUHK$01$}.
We compare the results with two evaluation settings (i.e., $100$ and $486$ test IDs) in Table~\ref{table:result_cuhk03}. For $486$ test IDs, the proposed method shows the best result. For $100$ test IDs, our method achieves the second best result, following \cite{journal/arxiv/geng16}. Note that \cite{journal/arxiv/geng16} fine-tuned the model which is learned from the CUHK$03$+Market$1501$, whereas we trained the model using $871$ training IDs of the CUHK$01$ dataset, following the settings in previous works~\cite{improved2015,chen2016deep,ChengGZWZ16,ZhaoLZW17}.

\vspace{.1cm}
\noindent\textbf{DukeMTMC}.
We follow the setting in \cite{journal/arxiv/lin17}
to conduct the experiments.
Table~\ref{table:result_duke} reports the results.
The proposed method achieves the best result for both with and without re-ranking.

\vspace{.1cm}
\noindent{\bf MARS}.
We also evaluate our method on one video-based person re-identification dataset~\cite{ZhengBSWSWT16}.
We use our approach to extract the representation
for each frame
and aggregate the representations of all the frames
using temporal average pooling,
which shows similar accuracy to other aggregation schemes (RNN and LSTM).
Table~\ref{table:result_mars} presents the comparison with the competing methods. Our method shows the highest accuracy over both image-based and video-based approaches.

\begin{table}[t]
\setlength{\tabcolsep}{5pt}
\centering
\caption{\small Accuracy comparison on DukeMTMC}\vspace{-.3cm}
\label{table:result_duke}
\resizebox{0.70\linewidth}{!}
{
\begin{tabular}{|l| |Sc|Sc|Sc|Sc|Sc|}
\hline
Rank & $1$ & $5$ & $10$ & $20$ & mAP\\
\hline\hline
Zheng et al.~\cite{conf/iccv/zheng17} & $67.7$ & -& -& -& $47.1$\\
Tong et al.~\cite{conf/cvpr/xiao17} & $68.1$ & -& -& -& -\\
Lin et al.~\cite{journal/arxiv/lin17} & $70.7$ & -& -& -& $51.9$\\
Schumann et al.~\cite{conf/cvprw/schumann17} & $72.6$ & -& -& -& $52.0$\\
Sun et al.~\cite{conf/iccv/sun17} & $76.7$ & $86.4$ & $89.9$& -& $56.8$\\
Chen et al.~\cite{conf/cvpr/chen17} (MobileNet) & $77.6$ & -& -& -& $58.6$\\
Chen et al.~\cite{conf/cvpr/chen17} (Inception-V3) & $79.2$ & -& -& -& $60.6$\\
Zhun et al.~\cite{journal/arxiv/zhong17} & $79.3$ &- & -& -& $62.4$\\
\quad\quad{\color{blue}+ re-ranking} & {\color{blue} $84.0$} & {\color{blue} -}& {\color{blue} -} & {\color{blue} -} & {\color{blue} $78.3$}\\
\hline
Proposed (Inception V1, OpenPose)& $82.1$ & $90.2$ & $92.7$ & $95.0$ & $64.2$\\
\quad\quad + dilation& $\mathbf{84.4}$ & $\mathbf{92.2}$ & $\mathbf{93.8}$ & $\mathbf{95.7}$ & $\mathbf{69.3}$\\
\quad\quad{{\color{blue}+ re-ranking}} & {\color{blue}$\mathbf{88.3}$} & {\color{blue}$\mathbf{93.1}$} & {\color{blue}$\mathbf{95.0}$} & {\color{blue}$\mathbf{96.1}$} & {\color{blue}$\mathbf{83.9}$} \\
\hline
\end{tabular}}\vspace{-.3cm}
\end{table}

\begin{table}[t]
\setlength{\tabcolsep}{5pt}
\centering
\caption{\small Accuracy comparison on MARS}\vspace{-.3cm}
\label{table:result_mars}
\resizebox{0.70\linewidth}{!}
{
\begin{tabular}[pos]{|Sl| |Sc|Sc|Sc|Sc|Sc|}
\hline
Rank & $1$ & $5$ & $10$ & $20$ & mAP\\
\hline\hline
Xu et al.~\cite{conf/iccv/xu17} (Video) & $44$ & $70$ & $74$ & $81$ & - \\
McLaughlin et al.~\cite{conf/cvpr/mclaughlin16} (Video) & $45$ & $65$ & $71$ & $78$ & $27.9$\\
Zheng et al.~\cite{ZhengBSWSWT16} (Video) & $68.3$ & $82.6$ & - & $89.4$ & $49.3$\\
Liu et al.~\cite{journal/arxiv/liu17} (Video) & $68.3$ & $81.4$ & - & $90.6$ & $52.9$\\
Zhou et al.~\cite{conf/cvpr/zhou17} & $70.6$ & $90.0$ & - & $97.6$ & $50.7$ \\
Li et al.~\cite{conf/cvpr/li17} & $71.8$ & $86.6$ & - & $93.1$ & $56.1$\\
\quad\quad{\color{blue}+ re-ranking} & {\color{blue}$83.0$} & {\color{blue}$93.7$} & {\color{blue}-} & {\color{blue}$97.6$} & {\color{blue}$66.4$}\\
Liu et al.~\cite{conf/cvpr/liu17} & $73.7$ & $84.9$ & - & $91.6$ & $51.7$\\
Hermans et al.~\cite{journal/arxiv/hermans17}& $79.8$ & $91.4$ & - & - & $67.7$\\
\quad\quad{\color{blue}+ re-ranking} & {\color{blue}$81.2$} & {\color{blue}$90.8$} & {\color{blue}-} & {\color{blue}-} & {\color{blue}$77.4$} \\
\hline
Proposed (Inception V1, OpenPose)  & $83.0$ & $92.8$ & $95$ & $96.8$ & $72.2$  \\
\quad\quad + dilation& $\mathbf{84.7}$ & $\mathbf{94.4}$ & $\mathbf{96.3}$ & $\mathbf{97.5}$ & $\mathbf{75.9}$ \\
\quad\quad{\color{blue}+ re-ranking} & {\color{blue}$\mathbf{85.1}$} & {\color{blue}$\mathbf{94.2}$} & {\color{blue}$\mathbf{96.1}$} & {\color{blue}$\mathbf{97.4}$} &  {\color{blue}$\mathbf{83.9}$} \\
\hline
\end{tabular}}\vspace{-.5cm}
\end{table}

\section{Conclusions}
We propose a new method for person re-identification.  The key factors that contribute to the superior performance of our approach
are as follows. (1) We adopt part maps where parts are not pre-defined but learned specially for person re-identification. They are learned to minimize the re-identification loss with the guidance of the pre-trained pose estimation model. (2) {\color{black}The part map representation provides a fine-grained/robust differentiation of the body part depending on their usefulness for re-identification.} (3) We use part-aligned representations to handle the body part misalignment problem. The resulting approach achieves superior/competitive person re-identification performances on the standard image and video benchmark datasets.

\appendix
\section{Appendix}

\subsection{Details of the Visualization}
\subsubsection{Figure 3} 
In Figure 3 (a) of the main manuscript, we visualize the appearance descriptors by mapping the normalized local appearance descriptors $\tilde{\mathbf{a}}_{xy}$ into 2D space using t-SNE~\cite{tsne}. Similarly, the normalized local part descriptors $\tilde{\mathbf{p}}_{xy}$ are visualized in Figure 3 (b).
\subsubsection{Figure 4 and 5} 
In Figures 4 and 5, the normalized feature maps are visualized following SIFTFlow~\cite{siftflow}. {\color{black} First, we collect the normalized local descriptors from all the test set images.} Then, we project the $c_A$(or $c_P$)-dimensional normalized local descriptor vector $\tilde{\mathbf{a}}_{xy}$(or $\tilde{\mathbf{p}}_{xy}$) onto the 3D RGB space, by mapping the top three principal components of descriptor to the RGB.

\subsection{Additional Visualization Examples of Feature Maps}
We show the additional visualization examples of feature maps on the MARS dataset in Figure~\ref{fig:vis_mars}.
For a given input image (left), appearance (center) and part (right) maps encode the appearance and body parts, respectively.
It shows how the appearance maps distinguish different persons while being invariant for each person. By contrast, the part maps encode the body parts independently from their appearance.

\setcounter{figure}{6}
\begin{figure}[t]
\centering
	\begin{minipage}{0.99\linewidth}
	\begin{minipage}{0.24\linewidth}
\centering
      \includegraphics[width=0.99\linewidth]{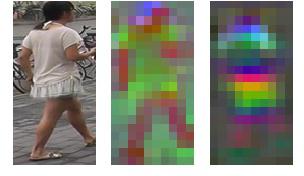}
      \includegraphics[width=0.99\linewidth]{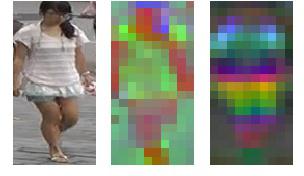}
      \includegraphics[width=0.99\linewidth]{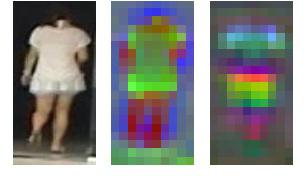}
      \includegraphics[width=0.99\linewidth]{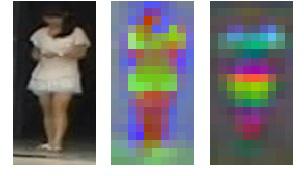}
    \end{minipage}
	\begin{minipage}{0.24\linewidth}
\centering
      \includegraphics[width=0.99\linewidth]{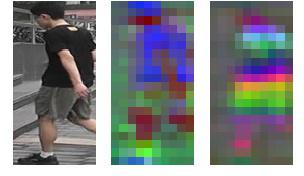}
      \includegraphics[width=0.99\linewidth]{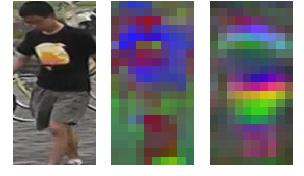}
      \includegraphics[width=0.99\linewidth]{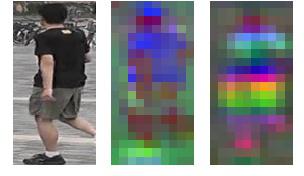}
      \includegraphics[width=0.99\linewidth]{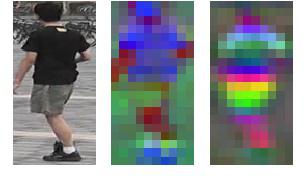}
    \end{minipage}
	\begin{minipage}{0.24\linewidth}
\centering
      \includegraphics[width=0.99\linewidth]{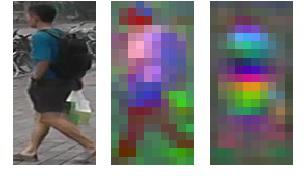}
      \includegraphics[width=0.99\linewidth]{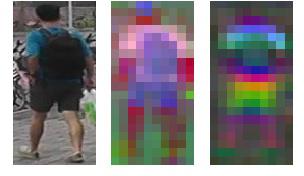}
      \includegraphics[width=0.99\linewidth]{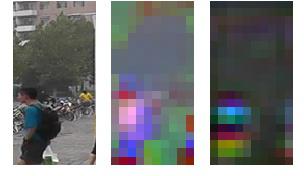}
      \includegraphics[width=0.99\linewidth]{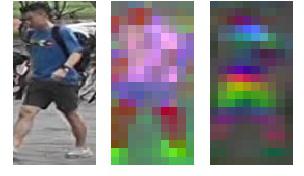}
    \end{minipage}
	\begin{minipage}{0.24\linewidth}
\centering
      \includegraphics[width=0.99\linewidth]{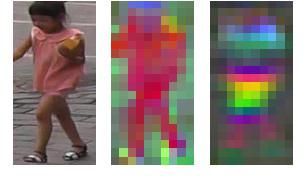}
      \includegraphics[width=0.99\linewidth]{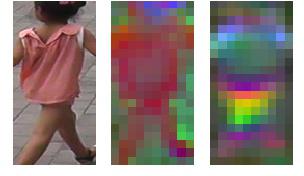}
      \includegraphics[width=0.99\linewidth]{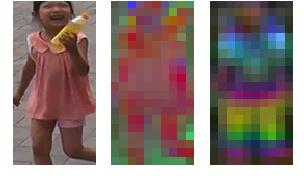}
      \includegraphics[width=0.99\linewidth]{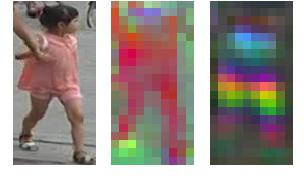}
    \end{minipage}

    \end{minipage}
\vspace{-.2cm}
\caption{Visualization of the appearance maps $\mathbf{A}$ and part maps $\mathbf{P}$ obtained from the proposed method on the MARS dataset. For a given input image (left), appearance (center) and part (right) maps encode the appearance and body parts, respectively.}
\label{fig:vis_mars}
\end{figure}

\setcounter{table}{6}
\begin{table}[b]
\setlength{\tabcolsep}{5pt}
\caption{\small Accuracy comparison of the baseline, proposed method, and its variation}\vspace{-.3cm}
\label{table:relu}
\centering
\scriptsize
\begin{tabular}[pos]{|Sl|Sl| |Sc|Sc|Sc|Sc|Sc|}
\hline
& Rank & $1$ & $5$ & $10$ & $20$ &mAP \\
\hline
\multirow{3}{*}{Market-$1501$} &
Baseline 				& $81.6$ 	& $92.0$ 	& $95.0$ 	& $96.9$ 	& $63.6$\\
&Proposed (original)		& $90.2$ 	& $96.1$ 	& $97.4$ 	& $98.4$ 	& $76.0$\\
&Proposed ({\it non-negative})		& $89.5$ 	& $95.6$ 	& $97.3$ 	& $98.1$ 	& $76.1$\\
\hline
\multirow{3}{*}{MARS} &
Baseline 				& $76.8$ 	& $89.8$ 	& $92.3$ 	& $94.6$ 	& $63.1$\\
&Proposed (original)	& $83.0$ 	& $92.8$ 	& $95.0$ 	& $96.8$ 	& $72.2$\\
&Proposed ({\it non-negative})		& $83.8$ 	& $94.3$ 	& $96.1$ 	& $97.2$ 	& $74.1$\\
\hline
\multirow{3}{*}{Duke} &
Baseline 				& $70.6$ 	& $83.8$ 	& $87.8$ 	& $91.2$ 	& $50.6$\\
&Proposed (original)	& $82.1$ 	& $90.2$ 	& $92.7$ 	& $95.0$ 	& $64.2$\\
&Proposed ({\it non-negative})	& $82.0$ 	& $90.6$ 	& $93.2$ 	& $95.2$ 	& $65.1$\\
\hline
\end{tabular}\vspace{-.6cm}
\end{table}

\subsection{Additional Analysis}

\subsubsection{Effect of non-negative part descriptors}
We test one variation of the proposed model, i.e., a model with non-negative part descriptors. In this model, we restrict the part descriptors $\mathbf{p}_{xy}$ to be element-wise  non-negative by adding a ReLU layer after the part map extractor $\mathcal{P}$. It causes the local part similarity to be always non-negative, and therefore the sign of the local similarity (Eq.9) depends only on the sign of the local appearance similarity. Table~\ref{table:relu} shows the results on the Market-$1501$, MARS, and Duke dataset. We use the same baseline used in Figures 6 (a) and (b). Overall, the proposed method and the non-negative variant show similar accuracy in terms of rank@1. The non-negative variant shows slightly improved accuracy in terms of mAP.

\subsection{Body Joints and Limbs}
Our model adopts the sub-network of the pose estimation network (OpenPose~\cite{conf/cvpr/cao17} $\mathcal{P}_{pose}$) to form the part map extractor $\mathcal{P}$, i.e., from the image input to the output of the stage2 ({\it concat\_stage3}). It generates a $185$-dimensional feature map which consists of a $19$-dimensional joint confidence map, $38$-dimensional limb confidence map, and $128$-dimensional internal feature map. Table~\ref{table:body_part} lists the body joints and limbs. For a further detailed representation, please refer to \cite{conf/cvpr/cao17}.

\begin{table}[t]
\fontsize{9}{9}\selectfont
\begin{center}
\centering
\caption{Joints and limbs used in OpenPose.
A limb refers to a connection of two joints.
}\vspace{-.3cm}
\label{table:body_part}
\begin{tabular}[pos]{|c |c|}
\hline
\multirow{2}{*}{joints} & nose, reye, leye, rear, lear, neck, rsho, lsho, relb, lelb, \\
& rwri, lwri, rheap, lheap, rkne, lkne, rank, lank, background\\ 
\hline
\multirow{4}{*}{limbs} & neck-lsho, neck-rsho, neck-lheap, neck-rheap,\\
& lsho-lelb, lelb-lwri, rsho-relb, relb-rwri,  \\
& lheap-lkne, lkne-lank, rheap-rkne, rkne-rank,\\
& nose-neck, nose-leye, leye-lear, nose-reye, reye-rear, \\
& lear-lsho, rear-rsho\\ 
\hline
\end{tabular}
 \vspace{-0.5cm}
\end{center}
\end{table}

\end{document}